%% file: paperICPM27.tex
\documentclass[manuscript]{acmart}

\usepackage{tikz}
\usetikzlibrary{arrows.meta,calc,fit,backgrounds,shapes.geometric,positioning,decorations.markings}
\usepackage[table]{xcolor}
\usepackage{tcolorbox}
\usepackage{bm} 

\usepackage{listings}
\newcommand{\GoalCat}{\textsf{GoalCat}}

\definecolor{csPurple}{RGB}{112,48,160}   
\newcommand{\coloredcircle}[2]{%
  \tikz[baseline=(char.base)]{
    \node[shape=circle, draw=none, fill=#1, inner sep=1pt, minimum size=10pt] (char)
    {\textcolor{white}{\small\sffamily\bfseries #2}};
  }%
}\newcommand{\cirtext}[1]{\coloredcircle{csPurple}{#1}}

\newcommand{\rqanswer}[2]{\begin{tcolorbox}[colframe=csPurple, colback=csPurple!15, title=\textbf{#1}]#2\end{tcolorbox}}

\AtBeginDocument{%
  }

\setcopyright{acmlicensed}
\copyrightyear{2027}
\acmYear{2027}
\acmDOI{XXXXXXX.XXXXXXX}
\acmConference[ICPM 2027]{ACM International Conference on Process Mining}{2027}{TBD}
\acmISBN{978-1-4503-XXXX-X/2027/XX}

\begin{document}

\title{Goal-driven Variant Categorization}

\author{Daniel Calegari}
\affiliation{%
  \institution{Universidad ORT Uruguay}
  \city{Montevideo}
  \country{Uruguay}
}
\email{calegari@ort.edu.uy}

\author{Daniel Amyot}
\affiliation{%
  \institution{University of Ottawa}
  \city{Ottawa}
  \country{Canada}
}
\email{damyot@uottawa.ca}

\begin{abstract}
Process discovery rarely yields a single coherent process structure. For analysis, a common step is to cluster process variants based on structural similarity and then assign business meaning to the resulting groups. Since these partitions are not derived from the organization’s goals, analysts must manually interpret and consolidate variants into business-meaningful categories. This judgment-intensive step becomes increasingly difficult as the number and complexity of variants grow.
In this paper, we propose a goal-driven approach to variant categorization that reverses this workflow. 
We first author an organization's goal model that predefines the categorization axis. Each variant is transformed into a textual narrative describing its behavior, and a Large Language Model (LLM) interprets it in the context of the goal model and assigns the variant to the most appropriate category. LLM-based semantic reasoning connects low-level process behavior with analyst-defined business goals.
We instantiate this approach end-to-end and evaluate it on three public logs differing substantially in scale and behavioral diversity. Goal-model guidance yields partitions that differ from those produced by unguided induction and respond to controlled edits to the declared alternatives, at the cost of authoring a goal model.
\end{abstract}



\maketitle

\section{Introduction}
\label{sec:introduction}

Process analysis on real-life event logs must reconcile two perspectives: the behavior recorded in execution data and the business meaning the organization assigns to it. Often, process discovery generates unreadable models, and the standard approach to joining the two perspectives is to \textit{cluster} trace variants by structural similarity~\citep{song2008traceclustering,amling2025bridging} and then attach business meaning (often as \textit{goals}) to express categories the organization recognizes. However, this is usually a judgment-heavy step, which becomes costly as the diversity and number of variants grow.

Recent work narrows this gap from both sides. On the clustering side, pairing a structural partition with a generated per-cluster description improves readability, yet the boundary is still induced from activity co-occurrence, and neither the boundary nor the label is checked against the process~\citep{amling2025bridging}. On the goal side, goal-oriented process mining supplies an external referent~\citep{ghasemi2020goalorientedreview}, but the closest approach decides only whether each variant satisfies a declared objective and sets the unsatisfying variants aside rather than categorizing them~\citep{ghasemi2025goped}. As far as we know, no existing work uses business semantics to define an explicitly declared categorization axis prior to assignment, nor does it treat uncategorized behavior as an inspectable coverage gap.

In this paper, we propose a goal-driven variant categorization approach that fixes the categorization axis in advance by declaring it in a goal model. A large language model (LLM) then derives a category taxonomy from that model and assigns trace variants to its categories.
The goal model is written in the standard Goal-oriented Requirement Language (GRL)~\citep{itu2018urn,amyot2022urnsurvey} and is set before any variant is assigned. Each variant is rendered as a narrative with the trace-textualization mechanism of LUPIN~\citep{pasquadibisceglie2024lupin}, and the LLM classifies each narrative into one of the taxonomy's categories. Narratives assigned to no category are reported as an inspectable residual, which may indicate a goal-model coverage gap. The approach then produces per-category descriptions, process models, and conformance measures for a human analyst to review, with the option to recategorize any variant. 

The contributions are: \cirtext{1}~a categorization method in which the goal model is the axis and the residual is a first-class output; \cirtext{2}~\GoalCat~\citep{GoalCat}, a publicly available Python implementation on the standard process-mining stack, with a guided mode and an open (goal-model-free) mode; and \cirtext{3}~a pre-registered evaluation on three public logs~\citep{deleoni2015rtfm,mannhardt2016sepsis,vandongen2019bpic}, with a replication package, answering two research questions:
\begin{itemize}
  \item \textbf{RQ1 (Divergence).} To what extent does a taxonomy induced from a declared goal model partition the variants of a log differently from a taxonomy induced from the narratives alone, beyond run-to-run variation?
  \item \textbf{RQ2 (Robustness).} When one declared goal model alternative is removed, merged, or added, does the categorization change only for the variants that realized it?
\end{itemize}

In this paper, Section~\ref{sec:background} covers background and related work, Section~\ref{sec:architecture} presents the \GoalCat{} pipeline, Section~\ref{sec:evaluation} reports the evaluation, Sections~\ref{sec:discussions} and~\ref{sec:threats} discuss results and threats to validity, and finally Section~\ref{sec:conclusion} concludes.

\section{Background and Related Work}
\label{sec:background}

\paragraph{Variant clustering and LLM-based textualization}
\label{sec:bg-clustering}
Trace clustering methods differ mainly in how traces are encoded before the partition is computed. Early work groups traces by control-flow similarity~\citep{song2008traceclustering}. Later methods encode behavior around each event, such as conserved subsequences and their context~\citep{bose2009context}, or learn representations from logs~\citep {dekoninck2018act2vec}. In all of them, the groups are formed based on structural or behavioral proximity, and a business label is assigned afterward~\citep{song2008traceclustering}. LLMs enter this line through natural-language renderings of logs, models, and their abstractions~\citep{BKA24}. Recent clustering work pairs an explainable partition with a generated per-cluster description~\citep{amling2025bridging}. This improves readability, but the model is only asked to \emph{name} a group that has already been formed; the boundary and the label are not checked against the process. We use textualization for the opposite purpose. LUPIN renders a trace as a natural-language narrative for next-activity prediction~\citep{pasquadibisceglie2024lupin}; we invoke the same renderer once per complete variant and let the LLM decide only a bounded membership question against alternatives declared before taxonomy induction and assignment.

\paragraph{Goal models and goal-oriented process mining}
\label{sec:bg-grl}
The User Requirements Notation (URN) is a standard language that combines goals and processes, and its goal sublanguage is the Goal-oriented Requirement Language (GRL)~\citep{itu2018urn}. A GRL model is a graph of \emph{intentional elements}, i.e., goals, softgoals, tasks, and resources, held by actors, and connected by decomposition, contribution, and dependency links~\citep{amyot2022urnsurvey}. Decomposition is typed: an AND-decomposition requires every subelement to be satisfied, whereas an OR- or XOR-decomposition names alternative subelements, each on its own a sufficient realization of the parent. URN has two decades of reported use~\citep{amyot2022urnsurvey}, with tool support in jUCMNav~\citep{roy2006jucmnav}.

We use one construct from this language: the OR/XOR-decomposition of a goal into alternative tasks. Before taxonomy induction and assignment, it declares a discrete set of ways the goal can be met, and its named elements serve as category anchors. Other GRL elements inform category meaning without introducing additional category anchors. The two operators differ: XOR alternatives are mutually exclusive, whereas OR alternatives may co-occur~\citep{itu2018urn}.

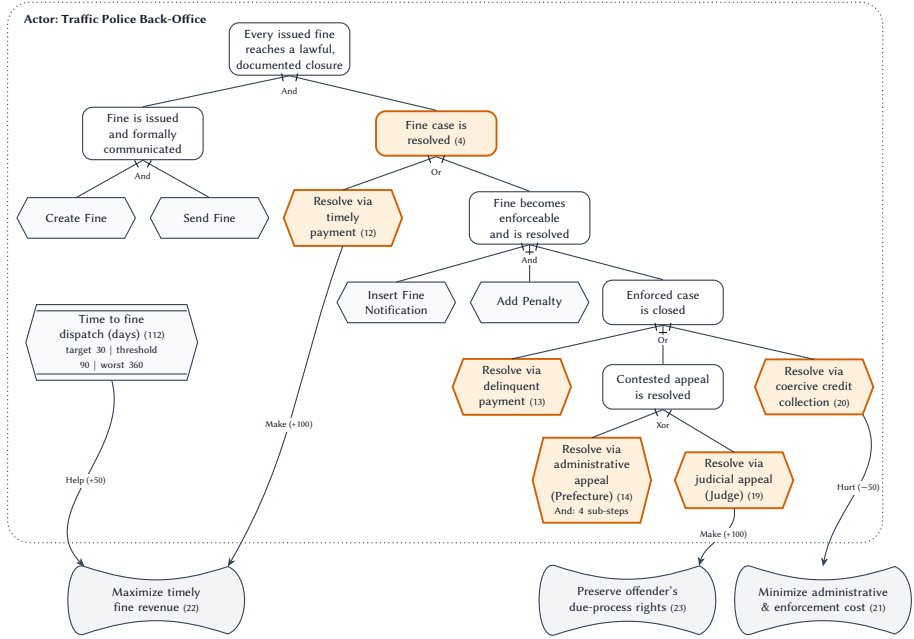
\begin{figure}[ht]
\centering
\resizebox{0.8\linewidth}{!}{%
  \input{figures/rtfm_goalmodel.tikz.tex}%
}
\caption{The RTFM goal model in GRL notation: rounded rectangles are goals, hexagons are tasks, the barred hexagon is a KPI, and the curved outlines are softgoals. Decomposition links carry a filled head at the decomposed parent and its And/Or/Xor operator; contribution links carry an open head and their negative/positive weight in the $[-100,100]$ interval.}
\Description{Goal model for the Road Traffic Fine Management process in GRL notation. One actor, the issuing police back-office, holds a goal tree AND-decomposed at the top into issuing a fine and resolving it, and OR/XOR-decomposed below into five ways a case can end: timely payment, delinquent payment, administrative appeal, judicial appeal, and coercive credit collection. Administrative appeal is AND-decomposed into four mandatory sub-steps. Three softgoals describe organizational priorities, and one KPI converts duration into a satisfaction level.}
\label{fig:rtfm-mini}
\end{figure}

Figure~\ref{fig:rtfm-mini} shows a GRL goal model for the Road-Traffic-Fine log (RTFM)~\citep{deleoni2015rtfm}. One actor, the traffic police back-office, holds a goal tree that is AND-decomposed at the top into issuing and resolving a fine, and OR/XOR-decomposed below into the ways a case can end. Three softgoals record what the organization wants to optimize while ending it. One of the model's four key performance indicators (KPIs) converts a measured duration into a satisfaction level that propagates along the contribution links~\citep{itu2018urn,fan2018arithmetic}. We developed the model for this work from publicly available sources (Section~\ref{sec:setup}).

Goal-oriented process mining connects such goal models to recorded behavior, so that the analysis refers to an explicit statement of the process's purpose~\citep{ghasemi2020goalorientedreview}. The closest approach, \textit{GoPED}, decides for each variant whether it satisfies a declared objective and sets the unsatisfying variants aside~\citep{ghasemi2025goped}. Two differences matter here. First, GoPED asks \emph{whether} declared goals are met (a filter), whereas we ask \emph{which} declared alternative a variant realizes (a categorization). Second, GoPED's set-aside variants are a filtered remainder; we make them a first-class residual that analysts can inspect.

\section{\GoalCat{} pipeline}
\label{sec:architecture}
\begin{figure}[t]
  \centering
  \resizebox{0.80\linewidth}{!}{%
    \input{figures/goalcat_pipeline.tikz.tex}%
  }
  \caption{The \GoalCat{} pipeline (phases I--IV). Solid arrows are sequence flows; dashed arrows are data associations, not step transitions.}
  \Description{The \GoalCat{} pipeline has four phases. Phase I extracts, profiles, textualizes, and samples trace variants. Phase II induces a taxonomy through either intent-guided or open induction. Phase III assigns every variant to a category or to the residual and analyzes the resulting categories. Phase IV supports analyst review and may return to taxonomy induction or goal-model revision.}
  \label{fig:goalcat-pipeline}
\end{figure}
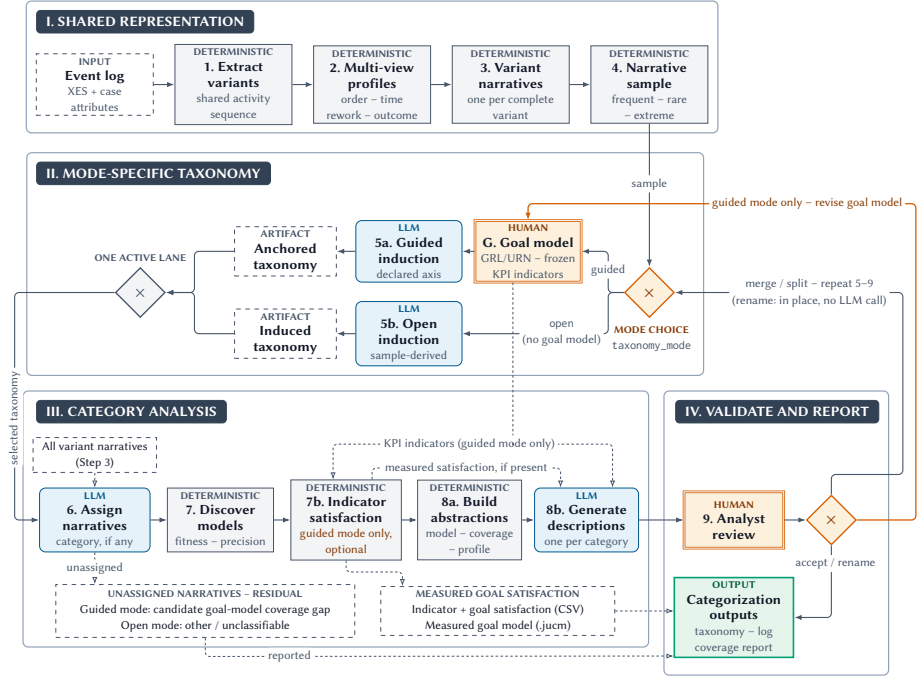

This section presents the goal-driven categorization method (contribution~\cirtext{1}) through the pipeline that implements it. \GoalCat{} runs trace-variant categorization in four phases (Figure~\ref{fig:goalcat-pipeline}) with two operating modes. \textbf{Phase~I} extracts the trace variants, profiles each variant across several behavioral views, and renders each variant as a natural-language narrative. \textbf{Phase~II} uses an explicit \texttt{taxonomy\_mode} switch (\texttt{intent\_guided} or \texttt{open}), defining whether a taxonomy is derived from an authored goal model or induced from narrative structure. \textbf{Phase~III} assigns narratives to the categories of the taxonomy, and describes quantitative and qualitative aspects of each category. Finally, \textbf{Phase~IV} submits those descriptions for analyst review and outputs the categorization results. In what follows, we detail each step.

\textbf{Goal model as declared taxonomy (Step G).} The pipeline requires a prior GRL/URN goal model with goals, actors, contribution links, and one declared \emph{axis root}: an OR- or XOR-decomposed goal whose alternative tasks form the categorization axis. We restrict the axis to OR/XOR points because any other basis would need an analyst-chosen criterion; other bases are future work. Naming the root fixes a single \emph{frontier} of candidate alternatives. The frontier is computed by descending from the root through nested OR/XOR points and stopping at any alternative that contains none, so an alternative that is AND-decomposed into mandatory sub-steps stays on the frontier as one alternative. A goal whose AND-decomposition joins several independent OR points declares no single axis, because a case realizes one alternative from each. The frontier later fixes the set the induced taxonomy must cover exactly (Step~5a) and the alternatives onto which each variant is projected (Step~6).
On RTFM, the descent yields the five frontier alternatives of Figure~\ref{fig:rtfm-mini}: timely payment, delinquent payment, administrative appeal, judicial appeal, and coercive credit collection. \emph{Administrative appeal} shows where the descent halts: it is AND-decomposed into four mandatory sub-steps and remains one alternative. The root decomposition is an inclusive OR, so one case may satisfy more than one alternative. The categories are disjoint only because Step~6 assigns a single label.

\textbf{Narrative representation (Steps 1--4).} These four steps build the narrative representation shared by both modes. Variant extraction groups log traces by their activity sequence. Profiling characterizes each variant along several views: activity order, duration and waiting times, rework, case outcome, resource, and the variant's frequency in the log, following the multi-perspective profiling proposed for trace clustering~\citep{song2008traceclustering}. Duration, rework, and resource distributions are aggregated across all cases of the variant. The event sequence rendered as text is not an aggregate; it comes from a single representative case, the one whose duration is closest to the variant's median, so the narrative reads as one coherent story. Textualization renders that profile as a natural-language narrative through LUPIN's event and trace templates~\citep{pasquadibisceglie2024lupin}.
Finally, a narrative sample (frequent, rare, and extreme variants) is drawn to calibrate the granularity at which taxonomy induction sets category boundaries. The assignment subsequently applies to every narrative, not just the sample.
Table~\ref{tab:rtfm-mini} summarizes eight RTFM narratives, each being rendered as a structured header (sampling reason, frequency, trace length, outcome), i.e., an aggregation over the variant's cases, followed by the activity sequence with inter-activity gaps. Nothing further about the variant is disclosed to the LLM. 

\begin{table}[t]
\centering
\small
\caption{Excerpt of RTFM narratives and the guided-arm category from the initial induction.}
\label{tab:rtfm-mini}
\rowcolors{2}{white}{csPurple!15}
\begin{tabular}{lll}
\toprule
\textbf{ID} & \textbf{Narrative} & \textbf{Guided category} \\
\midrule
V0004 & \texttt{Create Fine Payment (+1d)}                                              & timely payment \\
V0003 & \texttt{\dots\ Add penalty (+60d) Payment (+344d)}                              & delinquent payment \\
V0007 & \texttt{\dots\ Add penalty (+60d) Payment (+10d) Payment (+30d) \dots}          & delinquent payment \\
V0008 & \texttt{\dots\ Payment (+45d) Add penalty (+15d) Payment (+10d) \dots}          & delinquent payment \\
V0002 & \texttt{\dots\ Add penalty (+60d) Send for Credit Collection (+745d)}           & coercive collection \\
V0005 & \texttt{\dots\ Add penalty (+60d) Appeal to Judge (+21d)}                       & judicial appeal \\
V0006 & \texttt{\dots\ Send Appeal to Prefecture (+35d) \dots\ Receive Result \dots\ (+18d)}      & administrative appeal \\
V0001 & \texttt{Create Fine Send Fine (+134d)}                                          & --- (residual) \\
\bottomrule
\end{tabular}
\end{table}

\textbf{Taxonomy induction: intent-guided and open (Steps 5a/5b).} These modes differ in taxonomy induction and in the realization criterion used during assignment (Step~6).
In \texttt{intent\_guided} mode, the LLM receives the narrative sample together with the goal model and is asked to set category boundaries within the axis the model declares. It may not propose a different axis or introduce a category that is not traceable to a declared alternative. Both constraints are rechecked once the taxonomy is returned: a result that is not a partition of the declared axis invalidates the run, and milder anomalies are reported to the analyst. An \emph{empty} category, one to which no variant is later assigned, is not a defect; it is reported as declared-alternative coverage. In \texttt{open} mode, the LLM proposes categories directly from the narrative sample, an unsupervised analog of the clustering step in trace-clustering pipelines~\citep{song2008traceclustering,amling2025bridging}. Granularity is asymmetric: a category may anchor several alternatives under one inclusive-OR point, but two categories may not share an anchor, so induction cannot subdivide an alternative. Finer granularity has to be declared in the goal model. On RTFM, guided induction reproduces the five alternatives of Figure~\ref{fig:rtfm-mini} and the categories of Table~\ref{tab:rtfm-mini}. Each category is returned as a record with an anchor, evidence variants, and a description: \emph{judicial appeal} anchors to element~19, cites V0005, and its description restates the goal model's contribution links for that element (\emph{Make}~$(+100)$ towards due-process rights, \emph{Hurt}~$(-50)$ towards administrative cost); nothing in it is read off the log. Three payment variants that differ only in installment structure (V0003, V0007, V0008) all fall under \emph{delinquent payment} (Section~\ref{sec:threats}). Softgoals, contribution links, and KPIs inform category descriptions without introducing additional anchors; these descriptions may nevertheless influence Step~6's assignments.

\textbf{Assignment and residual (Step 6).} Assignment maps every narrative, not only the sample, to the category it best realizes, or to none. The assignment prompt receives category identifiers, names, and descriptions, along with batches of narratives, and requests exactly one category identifier or \texttt{null} per variant, along with a rationale. In guided mode, realization means satisfying the declared alternative; in open mode, it means matching the recurring pattern described by the category. Assignment records the LLM's rationale and a margin that the pipeline computes independently, based on a structural metric (edit distance over activity sequences) and a profile metric (outcome, duration, rework). Coverage is reported at two levels of granularity: variant-level coverage $C_V$ and case-weighted coverage $C_C$. They can differ substantially, since a long tail of low-frequency variants can depress $C_V$ with little effect on $C_C$. The residual means different things in the two modes. In \texttt{intent\_guided} mode, it is behavior not assigned to a declared alternative: a candidate coverage gap that may also reflect an unresolved case, insufficient narrative evidence, or assignment error. In \texttt{open} mode, there is no declaration to revise, so it is the analog of clustering noise. The limit of this reading is visible in Table~\ref{tab:rtfm-mini}: V0001, a still-open case, lies one edit from \emph{timely payment}, its nearest category, and only its rationale (\emph{ends with Send Fine and has no resolution}) distinguishes an open case from a genuine goal model coverage gap.

\textbf{Category analysis (Steps 7--8).} Per-category discovery applies the Inductive Miner~\citep{leemans2015inductiveminer} via PM4Py~\citep{BertiZS23} to each categorized subset independently, and a description step then renders the resulting evidence as prose judging the category's alignment with the model.

\textbf{Indicator satisfaction (Step~7b, guided mode only).} This optional step measures each KPI of the goal model over a category's own sublog, converts it to a satisfaction level, and propagates it through the goal model. The declared and the measured readings can disagree. On the full RTFM log, the goal model declares that judicial appeal \emph{Makes} (+100) the softgoal on due-process rights, yet across the five evaluated replicates the filing time with the Judge in that category is measured at 91 to 113 days against a statutory 30-day window, so that indicator scores $-100$ in every replicate. This step does not enter either research question; per-category indicator values are in the replication package.

\textbf{Analyst review (Step~9).} The analyst accepts, renames, merges, or splits categories through a decisions file. A merge or split requires re-running Steps~5--8. In \texttt{intent\_guided} mode, a recurrent coverage gap may be routed back to revise the goal model itself.

\paragraph{Implementation.}
\GoalCat{} (contribution~\cirtext{2}) is implemented in Python, built around PM4Py~\citep{BertiZS23}, and reusing the LUPIN module~\citep{pasquadibisceglie2024lupin} for textualization. Goal models are authored using the native interchange format of the GRL/URN environment jUCMNav~\citep{roy2006jucmnav}; KPI's conversion and propagation follow the standard's arithmetic and are checked against the closed-form equations in \citep{fan2018arithmetic}. Every LLM-backed step goes through a single adapter, so swapping the provider or model is a configuration change. 
The source code, including a web-based analyst interface, is publicly available~\citep{GoalCat}.

\section{Evaluation}
\label{sec:evaluation}

This section reports the evaluation (contribution~\cirtext{3}) of the two questions stated in the introduction, RQ1 (divergence) and RQ2 (robustness). The use of AI in conducting it is described in Section~\ref{sec:setup}. All the material is publicly available~\citep{GoalCat}.

\subsection{Setup}
\label{sec:setup}
\textit{Protocol.} Every run follows a protocol frozen before execution: one LLM (\texttt{gemini-3.5-flash-lite}, temperature 0), one narrative sample per log shared by all conditions, and one assignment batch size (25 narratives). Steps 7--9 were not executed in any evaluated condition: the comparison is defined on the partitions produced by Step~6, before category analysis or analyst revision. A pre-registered stability check reruns Step~5a five times per axis and requires an invariant anchor set; all four evaluated axes passed.

\textit{Logs and goal models.} Three public logs span a fifty-fold range in variant count: RTFM~\citep{deleoni2015rtfm} is high-volume and low-variety, Sepsis~\citep{mannhardt2016sepsis} is the opposite, and BPIC~2019~\citep{vandongen2019bpic} is high on both. Each log's goal model was authored for this study from publicly available information, without domain-expert review. Sepsis's root goal joins two independent OR frontiers under an AND decomposition, so it is treated as two axes, each with its own runs.

\textit{Arms.} Each axis is compared across five treatments: intent-guided ($T^G$, Step~5a with the goal model), open ($T^O$, Step~5b on the identical sample without it), \texttt{guided\_no\_sample}, and two label-list controls (\texttt{label\_list}, \texttt{label\_list\_strict}). The controls skip Step~5a and assign against the same externally authored list under the open and guided assignment criteria, respectively. List cardinality matches $T^G$ except for Sepsis admission, whose guided taxonomy has two categories whereas the shared Sepsis control list has five. Open and label-list runs are shared between Sepsis's two axis comparisons.

\textit{Replicates and statistics.} The frozen protocol fixed two replicates per arm. Guided and open were later extended to five replicates on RTFM and Sepsis and three on BPIC~2019, so that both arms' replicate-agreement floors rest on the same number of reruns. We report every completed replicate and state the $n$ for each table. Replicates are identical-input reruns of a temperature-0 endpoint, so we report means with descriptive min--max ranges and apply no inferential test. RQ2 is scoped to RTFM and Sepsis; BPIC~2019's perturbations were not run for cost reasons (Section~\ref{sec:feasibility}).

\textit{Use of AI in this research and evaluation.} An LLM coding assistant, Claude Code with Anthropic's Claude models (Opus~5 and Sonnet~5), was used throughout the research lifecycle, with the authors making every design decision and reviewing every artifact. Specifically, it (i)~generated most of the \GoalCat{} source code and its test suite from the authors' specifications; the KPI conversion and propagation code was additionally checked against the closed-form equations of~\citep{fan2018arithmetic}; (ii)~produced first drafts of the three goal models and of the RQ2 perturbed variants from the public sources cited in the replication package, which the authors revised (element naming, decomposition operators, and indicator value sets) and froze before any run; no domain expert reviewed them (Section~\ref{sec:threats}); (iii)~drafted the pre-registered protocol, the condition configurations, and the analysis scripts that compute coverage, partition agreement, and the reassignment rates, which the authors reviewed and which are guarded by regression tests; (iv)~generated the TikZ sources of Figures~\ref{fig:rtfm-mini} and~\ref{fig:goalcat-pipeline} from the authors' sketches; and (v)~assembled the replication package. Every categorization result reported below was produced by \texttt{gemini-3.5-flash-lite} inside the pipeline, and every number in the tables was computed by the deterministic analysis scripts from the archived run outputs, not by the assistant.

Table~\ref{tab:setup} summarizes the evaluation logs and the intent-guided partition ($T^G$). $|T^G|$ is the number of categories in the guided taxonomy (Step~5a); it is invariant across every replicate of every axis and is therefore reported without a range. Residual (var.) and Residual (cases) are the percentages of variants and cases, respectively, assigned to no category (Step~6), given as mean [min--max] over the $n$ guided replicates because Step~6 is not deterministic; a bare residual value denotes a range whose endpoints coincide at the displayed precision.

\begin{table}[b]
\centering
\caption{Evaluation logs and the intent-guided partition ($T^G$).}
\label{tab:setup}
\rowcolors{2}{white}{csPurple!15}
\begin{tabular}{llrrrrrr}
\toprule
\textbf{Log} & \textbf{Axis} & \textbf{Variants} & \textbf{Cases} & $\bm{|T^G|}$ & $\bm{n}$ & \textbf{Residual (var.)} & \textbf{Residual (cases)} \\
\midrule
RTFM      & resolution      & 231      & 150{,}370 & 5 & 5 & 2.8\% [0.9--7.4]   & 13.8\% \\
Sepsis    & admission       & 846      & 1{,}050   & 2 & 5 & 10.4\%             & 22.9\% \\
Sepsis    & discharge       & 846      & 1{,}050   & 5 & 5 & 18.2\% [17.1--21.0] & 29.6\% [28.6--32.1] \\
BPIC 2019 & matching regime & 11{,}973 & 251{,}734 & 4 & 3 & 20.7\% [19.3--21.6] & 16.3\% [16.0--16.8] \\
\bottomrule
\end{tabular}
\end{table}

\subsection{RQ1 (divergence): intent-guided versus open induction}

Both arms categorize the same variants of the same log with the same model and narrative sample. They differ in the origin of their taxonomies and, in the evaluated implementation, in a mode-specific realization criterion in the Step~6 prompt. The comparison therefore contrasts two complete treatments; it does not isolate taxonomy provenance alone. We compare coverage, use of the declared alternatives, and the resulting partitions.

\textit{Coverage.} The pattern differs across logs. On Sepsis, the guided residual is larger than the open residual on both axes. On RTFM, case-weighted residual rates are nearly identical because one unresolved high-frequency variant dominates both, although variant-weighted rates differ. In BPIC~2019, the residual ranges overlap under both weightings: the guided mean is lower by variant count but higher by case count. Coverage therefore describes assignment under a particular declared frame, narrative representation, and assignment mechanism; it is not a quality score.

Guided's residual is stable across BPIC~2019's reruns, whereas open's swings by roughly 55 percentage points (variant-weighted) across identical reruns, ten times guided's widest swing anywhere in the study (RTFM, under 7 points). The open arm's category names remain stable across reruns; the portion of the log they cover does not. On the largest and most diverse log, open's coverage is thus unpredictable from one identical rerun to the next, which the frozen two-replicate protocol could register as a swing but not distinguish from a single outlier.

\textit{Declared alternatives.} No declared alternative goes unused. Every alternative on the frontier is realized by at least one variant on every axis and in every replicate: 5/5 on RTFM, 2/2 on Sepsis's admission axis, 5/5 on its discharge axis, and 4/4 on BPIC~2019. Several alternatives are nonetheless near-vanished in full scope, and per-category figures computed from them are anecdotal.

\textit{Partition agreement.} The two arms do not cut the log along the same lines. We measure the agreement between two categorizations of the same variants with Adjusted Mutual Information (AMI), which is 1 for identical partitions and about 0 for unrelated ones, treating the residual as a block of its own (Table~\ref{tab:divergence}). Guided-vs-open agreement averages 0.17 to 0.30 across the four axes (variant-weighted). Each arm's agreement with its own reruns is the noise floor for that comparison. Guided's minimum within-arm agreement exceeds the maximum guided-vs-open agreement on every axis and weighting, separating the observed cross-arm differences from the variation observed among guided reruns. Open's floor is less consistent: on RTFM (both weightings) and on BPIC~2019 case-weighted, it overlaps the guided-vs-open band, so on those three readings the conclusion rests on guided's stability alone. On RTFM, hand inspection shows a systematic split-and-merge rather than scattered noise: the open arm folds the two appeal categories into one, splits the payment categories by execution detail, and sends most of the guided residual into the merged appeal category (contingency counts are in the replication package~\citep{GoalCat}).

\begin{table}[b]
\centering
\footnotesize
\caption{Partition agreement (AMI, residual as its own block), mean [min--max] over replicate pairs. Each arm's own agreement is the noise floor for the guided-vs-open column.}
\label{tab:divergence}
\rowcolors{2}{white}{csPurple!15}
\begin{tabular}{llcccccc}
\toprule
 & & \multicolumn{3}{c}{\textbf{Variant-weighted}} & \multicolumn{3}{c}{\textbf{Case-weighted}} \\
\cmidrule(lr){3-5}\cmidrule(lr){6-8}
\textbf{Log} & \textbf{Axis} & \textbf{Guided-guided} & \textbf{Open-open} & \textbf{Guided vs.\ open} & \textbf{Guided-guided} & \textbf{Open-open} & \textbf{Guided vs.\ open} \\
\midrule
RTFM      & resolution      & 0.58 [0.49--0.69] & 0.50 [0.42--0.65] & 0.26 [0.15--0.46] & 0.99 [0.98--1.00] & 0.93 [0.87--1.00] & 0.91 [0.87--0.96] \\
Sepsis    & admission       & 0.88 [0.84--0.92] & 0.51 [0.43--0.56] & 0.30 [0.27--0.32] & 0.92 [0.89--0.94] & 0.61 [0.55--0.66] & 0.44 [0.40--0.46] \\
Sepsis    & discharge       & 0.76 [0.68--0.86] & 0.51 [0.43--0.56] & 0.25 [0.21--0.32] & 0.79 [0.72--0.88] & 0.61 [0.55--0.66] & 0.36 [0.30--0.43] \\
BPIC 2019 & match. regime & 0.51 [0.48--0.54] & 0.40 [0.36--0.43] & 0.17 [0.17--0.18] & 0.69 [0.61--0.83] & 0.54 [0.45--0.70] & 0.40 [0.27--0.49] \\
\bottomrule
\end{tabular}
\end{table}

\textit{Controls.} The two label-list controls replace the induced taxonomy with a fixed, externally authored list of the same size (except on Sepsis admission). Agreement between guided and either control stays well below guided's own replicate floor on every axis, so neither the number of categories nor the assignment criterion explains the divergence on its own. Each list was deliberately drawn along a different axis from the goal model, so the controls cannot separate the goal model's axis from its vocabulary. Full results are in the replication package~\citep{GoalCat}; Section~\ref{sec:threats} discusses the caveats.

\textit{Baselines.} Structural clustering over activity-presence vectors (no order or timing; HDBSCAN) over-segments every log relative to the declared axis, and none of its three partitions aligns with either LLM partition. A five-rule classifier written once for RTFM from the process description, without tuning against the log, uses activity presence and the order of \texttt{Add penalty} and \texttt{Payment}. Its fixed priority assigns appeals before collection or payment, even when a later activity closes the case. It agrees with the guided assignment for approximately 97\% of cases in replicate~1, with agreement above 99\% in four of the five guided categories. Disagreement concentrates in the delinquent payment category.

\rqanswer{Answer to RQ1 (Divergence)}{Across all four axes and both weightings, cross-arm agreement remains below the observed guided--guided agreement range; mean variant-weighted AMI is 0.17--0.30. Separation from open--open agreement is incomplete on RTFM under both weightings and on case-weighted BPIC~2019. Every declared alternative is used, while residual size depends on the log and weighting. Neither tested label-list control reproduces the guided partition, but the comparison does not isolate the factors in which the treatments differ.}

\subsection{RQ2 (robustness): goal-model perturbations}

Here we edit the goal model itself and re-run the guided arm five times for each edited model. Each axis is perturbed in three ways: (A)~one declared alternative is removed; (B)~two are merged (refused where a decomposition point would be left with a single child); and (C)~one plausible but unrealized alternative is added under the axis, grounded in the same public material as the base model (e.g., RTFM's \emph{fine annulment}). Table~\ref{tab:rq2} reports two rates for A and B, each representing the mean over five replicates, with its min--max range. \textit{TargetReassignment} is the share of variants realizing the perturbed alternative(s) that change category. It checks that the edit took effect: it is exactly 100\% for every removal and falls slightly short for both merges (97.5\% RTFM, 99.3\% Sepsis discharge), because the merge tool retains one target's original ID. \textit{CollateralReassignment}, the share of \emph{every other} variant that changes category, is the quantity of interest. For perturbation~C, no baseline variant realizes the added alternative, so we report \textit{DistractorUptake}, the share of variants assigned to it. Step~5a admits the distractor into the taxonomy in all 15 replicates, and Step~6 assigns it zero variants in every one.

Collateral movement means little on its own, since two identical reruns already reassign some variants. Hence, every rate is read against $C^\emptyset$, the same rate measured between pairs of five unperturbed guided replicates within one fixed collateral population. Those ten pairs are not independent, so $C^\emptyset$ is a descriptive min--max range rather than a null distribution. Of the eight conditions (three axes times three edits, less the refused Sepsis admission merge), four have ranges that sit entirely inside their $C^\emptyset$ range: Sepsis admission's removal sits entirely below it, quieter than ordinary replicate noise, and three conditions (Sepsis admission's and discharge's distractors, and Sepsis discharge's removal of \emph{Release E}) extend a few points past its upper edge. In these three conditions, the five-replicate mean remains within the baseline range, but some replicates exceed its upper edge. This descriptive overlap does not establish equivalence or rule out the possibility of an additional perturbation effect.

\begin{table}[ht]
\centering
\small
\caption{Goal-model perturbations, guided arm, mean [min--max] over five replicates per condition (Sepsis admission has no merge: OR point with only two children). $n$ in the Target column is the number of variants realizing the perturbed alternative(s) (n/a for C); $n$ in the Collateral column is the rest. $^\dagger$C's Target column reports DistractorUptake in place of TargetReassignment.}
\label{tab:rq2}
\rowcolors{2}{white}{csPurple!15}
\begin{tabular}{lllrrr}
\toprule
\textbf{Log} & \textbf{Axis} & \textbf{Edit} & \textbf{Target rate ($\bm{n}$)} & \textbf{Collateral rate ($\bm{n}$)} & $\bm{C^\emptyset}$ \textbf{range} \\
\midrule
RTFM   & resolution & A: remove \emph{coercive collection}     & 100.0\% (34)                    & 20.5\% [17.8--24.9] (197) & 15.7--39.6\% \\
RTFM   & resolution & B: merge \emph{delinquent}+\emph{coercive}   & 97.5\% [93.8--100] (81)          & 20.7\% [16.7--23.3] (150) & 16.0--43.3\% \\
RTFM   & resolution & C: add \emph{fine annulment}$^\dagger$    & 0.0\% [0.0--0.0] uptake (n/a)    & 19.7\% [19.0--21.6] (231) & 14.3--35.1\% \\
Sepsis & admission  & A: remove \emph{Admission IC}             & 100.0\% (89)                    & 0.05\% [0.00--0.13] (757) & 0.4--2.4\%   \\
Sepsis & admission  & C: add \emph{High-Dependency Unit}$^\dagger$ & 0.0\% [0.0--0.0] uptake (n/a) & 2.2\% [1.4--3.2] (846)    & 1.2--2.7\%   \\
Sepsis & discharge  & A: remove \emph{Release E}                & 100.0\% (6)                     & 8.9\% [6.9--11.7] (840)   & 3.2--9.3\%   \\
Sepsis & discharge  & B: merge \emph{Release A}+\emph{B}         & 99.3\% [97.5--100] (640)        & 3.5\% [3.4--3.9] (206)    & 2.9--15.5\%  \\
Sepsis & discharge  & C: add \emph{Release F}$^\dagger$          & 0.0\% [0.0--0.0] uptake (n/a)   & 8.9\% [7.6--10.0] (846)   & 3.2--9.2\%   \\
\bottomrule
\end{tabular}
\end{table}

\rqanswer{Answer to RQ2 (Robustness)}{Largely yes. Collateral reassignment stays within or slightly above the range observed among unperturbed replicates in all eight conditions. An added alternative that no variant realizes is admitted into the taxonomy in all 15 runs but receives no variant in any of them. Target reassignment (100\% for removals and 97.5--99.3\% for merges) confirms that each edit took effect. On RTFM, the unperturbed range is wide, so its three conditions can detect only a large reorganization.}

\subsection{Feasibility}
\label{sec:feasibility}

Across all runs in this study, the pipeline issued roughly 6,900 hosted-LLM calls (58.8M input tokens, 9.9M output tokens) at an estimated cost of USD 42.3 on the gemini-3.5-flash-lite paid endpoint, running on a single laptop-class workstation (8-core Apple M3, 8 GB RAM). A single guided or open condition costs roughly USD 0.05 on RTFM, USD 0.17 on Sepsis, and USD 3.09 on BPIC 2019. Cost tracks variant count, since assignment issues one call per batch of 25 narratives (10, 34, and 479 calls, respectively), while Step 5a is a single call regardless of log size. Wall-clock time does not follow cost: the deterministic steps dominate a guided condition's wall-clock time on the small logs (81–87\% on RTFM), whereas on BPIC 2019, the LLM assignment step alone runs for 600–2,100 s. A laptop-class local model (qwen2.5:3b via Ollama) was not a viable substitute: the frozen batch size exceeded its context window. These figures are indicative, not a benchmark; a full resource characterization accompanies the replication package.

\section{Discussion}
\label{sec:discussions}

\textbf{Guided and open treatments produce different partitions} (RQ1). Cross-arm agreement falls below the observed guided--guided range on every axis and weighting, but it overlaps the open--open range on RTFM and case-weighted BPIC~2019. RTFM inspection identifies a split-and-merge pattern, while its mean case-weighted cross-arm AMI of 0.91 shows that the difference is much smaller for frequent cases. This comparison describes the complete treatments and does not isolate the effect of goal-model guidance.

\textbf{Controlled goal-model perturbations provide the strongest evidence about the mechanism} (RQ2). Collateral reassignment is generally comparable to observed unperturbed variation, although three conditions partly exceed its range. The result supports sensitivity to declared alternatives with limited additional disruption relative to baseline variation; it does not imply negligible absolute movement or validate the categories. Distractors receive no assignments.

\textbf{The declared frame provides an explicit reference for inspecting the residual.} The unresolved RTFM variant carrying 13.6\% of cases remains residual under both arms in every replicate; the guided model relates its exclusion to the declared resolution alternatives. Guided residual rates are more stable than open ones on Sepsis and BPIC~2019, but the guided residual is not uniformly larger. Coverage depends on the declared frame, narrative representation, and assignment mechanism; it does not assess the quality of categorization or illustrate gaps in the goal model coverage.

\textbf{On a log with a small, well-separated outcome vocabulary, the goal model contributes traceability rather than discriminative power}: a hand-written classifier matches the guided partition almost exactly, whereas structural clustering fails, because three payment variants differing only in installment structure collapse to one activity-occurrence vector. The guided arm also places those three variants in one category for a different reason: the goal model declares a single alternative there. \textit{The partition's granularity is set by what the model declares, not by what the log distinguishes.} That ceiling belongs to the construct studied here, one OR/XOR frontier projected to a single label, not to fixing the axis in advance, which admits finer-grained declared bases.

The goal model fixes the taxonomy's structure. Across all 46 guided and ablation inductions on the four axes, exactly one anchor set occurs per axis, with and without the narrative sample. Generated descriptions and rationales vary; the category descriptions feed Step~6, so assignment variation may reflect both description changes and Step~6's own nondeterminism. That $|T^G|$ equals the frontier size everywhere is enforced on RTFM and Sepsis (two categories may not share an anchor, and the merge rule never fired) but was \emph{observed} on BPIC~2019, where seven guided runs executed before the partition check existed recovered the declared frontier exactly ($7/7$) from the goal-model excerpt alone.

Step~7b (guided arm only) was pre-registered for RTFM alone, the only log whose indicators are organizational targets rather than percentiles computed from the log itself. Because its sublogs are Step~6's output, \textbf{indicator satisfaction characterizes the partition rather than validating it}.

Folding interleavings of concurrent activities into partial-order classes would not cut assignment cost where it matters: RTFM's 231 variants fold to 85, but BPIC~2019's 11{,}973 fold only to 11{,}463, because the variant explosion in that log is rework-driven rather than parallelism-driven.

\section{Threats to Validity}
\label{sec:threats}

\paragraph{Conclusion validity.} The evaluation covers three logs and two to five replicates per condition, providing limited evidence about variability and generalization. Guided and open replicate counts are matched within each axis, but differ across logs and from the controls. Pairwise comparisons share runs, and min--max ranges depend on the number of observations; these ranges are descriptive and do not establish statistical significance or equivalence.
Reliability is controlled by construction. We record and hash every condition's inputs, and a paired comparison that departs from its design is re-run and not reported. Step~5a reproduces the same anchor set in every guided rerun of every axis, whereas Step~6 assignments vary; the observed guided--guided AMI ranges exceed the corresponding open--open ranges in Table~\ref{tab:divergence}. Cost and latency were measured on one workstation against a hosted endpoint whose behavior varied during the measurement window; they show the pipeline is affordable at these scales and are not a benchmark.

\paragraph{Internal validity.} Perturbation~B merges two alternatives: it moves a category boundary and also lowers the taxonomy's cardinality, and the two effects cannot be separated here. The batch size was fixed for wall-clock/reliability reasons rather than varied, so it cannot explain within-pair differences, but its interaction with assignment quality is uncharacterized. Both are engineering choices rather than variables under study.

\paragraph{Construct validity.} No domain authority reviewed the goal models, the resulting categories, residuals, or the generated descriptions. Two claims are therefore out of reach: that the declared alternatives align with how the organization thinks about the process, and that a residual variant indicates a real gap rather than an omission by the model's author. The construct is also coarser than the log, as the RTFM payment variants of Section~\ref{sec:discussions} show. The single-label projection is a further limitation: an inclusive OR point permits several alternatives to hold at once, whereas assignment returns exactly one, and we did not measure how often that choice discards a second, genuinely realized alternative. On RTFM, the effect is measurable and concentrated as 120 variants lodge an appeal and then close by payment or collection, and the five identical-input guided replicates resolve between 42 and 74 of them by the appeal, the rest by the closing activity. The guided arm's variant-level instability is thus largest where the declared axis leaves the single-label choice open, which is an argument for declaring the tie-break in the goal model rather than leaving it to the LLM. Finally, the two arms differ in four respects: the guided arm is given an external axis, a fixed number of categories, the goal model's own vocabulary, and a stricter assignment criterion. The last of these plausibly shifts the residual boundary on its own, and none of the four is separable from these runs. The evaluation therefore characterizes the categorization mechanism, while the organizational relevance of its outputs and their usefulness to analysts remain unvalidated.

\paragraph{External validity.} Every run uses one LLM. The paired design only requires that both arms of a pair share it, so the comparison is sound, but we cannot say whether the divergence holds for other model families. The approach also assumes a goal model that declares one usable OR/XOR axis. Sepsis does not because its root joins two independent OR frontiers, so we split it into two axes. A model whose decomposition types are left unset silently defaults to AND and declares no axis at all. Finally, the authors used Step~9 only as a mechanism check; no domain expert reviewed a description, so the analyst-review loop runs but is not evaluated.

\section{Conclusion and Future Work}
\label{sec:conclusion}

We presented goal-driven variant categorization~\cirtext{1}, an approach that uses a GRL/URN goal model to induce a classification taxonomy and categorize trace variants. The goal model declares the alternative ways a goal is met, and those declared elements serve as category anchors, while behavior that matches none of them is reported as a residual, i.e., an inspectable candidate goal-model coverage gap. We instantiated the approach as \GoalCat~\cirtext{2}, a publicly available pipeline, and evaluated it on three public logs spanning a fifty-fold range in variant count~\citep{GoalCat}~\cirtext{3}.

On RQ1, the guided and open treatments partition the same variants differently. Cross-arm
agreement stays below the guided replicate range on every axis, but overlaps the open range on RTFM and on BPIC~2019 case-weighted. The arms also differ in four respects that this design cannot separate. On Sepsis and BPIC~2019, guided residuals are more stable than open ones, though not uniformly larger. For RQ2, the main result is that removing or merging a declared alternative moves the variants that realized it. Collateral reassignment stays generally within the range of unperturbed reruns, and an added alternative that no variant realizes receives none. Two findings bound the claim. On RTFM, a handwritten classifier agrees with the guided partition in 97\% of cases, but only in about two-thirds of variants. The goal model fixes the anchored structure at Step~5a, while the generated descriptions and Step~6 remain sources of variation. That remaining variation is concentrated where the declared axis leaves the single-label choice open (Section~\ref{sec:discussions}), so a goal model used as a categorization axis should also declare which alternative prevails when a case realizes several. In addition, at these scales, a run costs from a few cents to a few dollars per condition.

Several research directions follow. First, the goal models and the categorizations they produce require domain-expert validation, which would also settle whether a recurring residual marks a genuine gap in organizational intent. Second, other declared categorization axes (AND-decomposition phases, softgoal profiles, KPI bands) deserve to be instantiated, each with its own validity treatment. Third, how the approach degrades under partial or under-specified goal models deserves direct study. Fourth, extending Step~7b with the whether-question of goal-oriented process mining would give a satisfaction signal for the residual and a finer read within each category. Replication with other model families and an evaluation of the analyst-review loop itself remain open.

\section*{Data Availability}
The source code, experimentation scripts, and results are publicly available~\citep{GoalCat}.

\section*{Ethics and Privacy Statement}
Event logs are public: Road Traffic Fine Management~\citep{deleoni2015rtfm}, Sepsis
\citep{mannhardt2016sepsis}, and BPI Challenge 2019~\citep{vandongen2019bpic}. Goal models were authored from publicly available information. No institutional review was required.

\begin{acks}
\textbf{Use of AI.} GenAI tools were used in two distinct ways in this work, and the authors take full responsibility for all content. First, in the conduct of the research: as detailed in Section~\ref{sec:setup} (\emph{Use of AI in this research and evaluation}), Claude Code with Anthropic's models generated most of the \GoalCat{} source code and tests (Section~\ref{sec:architecture}, \textit{Implementation}), drafted the goal models and their RQ2 perturbations, drafted the pre-registered protocol and the analysis scripts, generated the TikZ sources of Figures~\ref{fig:rtfm-mini} and~\ref{fig:goalcat-pipeline}, and assembled the replication package~\citep{GoalCat}. Every such artifact was specified, reviewed, and revised by the authors, and the goal models were frozen before any run. The LLM under study, \texttt{gemini-3.5-flash-lite}, produced every categorization result reported in Section~\ref{sec:evaluation}, as the object of the experiments rather than as an assistant. Second, in writing: GenAI tools assisted with drafting, restructuring, and language editing of this manuscript. All claims, numbers, and interpretations were verified by the authors against the archived run outputs.
\end{acks}

\bibliographystyle{ACM-Reference-Format}
\bibliography{references}

\end{document}

%% file: figures/rtfm_goalmodel.tikz.tex
%
%

\begingroup%
\definecolor{GMInk}{HTML}{111827}%
\definecolor{GMLine}{HTML}{334155}%
\definecolor{GMFill}{HTML}{F8FAFC}%
\definecolor{GMAxis}{HTML}{D55E00}%
\definecolor{GMAxisFill}{HTML}{FDF0DD}%
\definecolor{GMSoftFill}{HTML}{F3F4F6}%

\hyphenpenalty=10000\relax%
\exhyphenpenalty=10000\relax%
\newcommand{\gmId}[1]{{\fontsize{6.4}{7}\selectfont (#1)}}%
\newcommand{\gmSub}[1]{{\fontsize{6.6}{7.4}\selectfont #1}}%

\begin{tikzpicture}[
  x=1cm, y=1cm,
  font=\sffamily\small,
  text=GMInk,
  goal/.style={
    draw=GMLine, line width=0.7pt, fill=white, rounded corners=2.2mm,
    align=center, inner xsep=3pt, inner ysep=4pt, text width=25mm, minimum height=10mm},
  axisroot/.style={
    goal, draw=GMAxis, line width=1.2pt, fill=GMAxisFill},
  hexbase/.style={
    draw=none, align=center, inner xsep=1pt, inner ysep=4pt,
    text width=21mm, minimum height=9mm},
  task/.style={hexbase,
    append after command={\pgfextra{\begin{pgfonlayer}{background}
      \path[draw=GMLine, line width=0.7pt, fill=GMFill]
        (\tikzlastnode.north west) -- (\tikzlastnode.north east)
        -- ([xshift=2.4mm]\tikzlastnode.east) -- (\tikzlastnode.south east)
        -- (\tikzlastnode.south west) -- ([xshift=-2.4mm]\tikzlastnode.west) -- cycle;
    \end{pgfonlayer}}}},
  anchortask/.style={hexbase,
    append after command={\pgfextra{\begin{pgfonlayer}{background}
      \path[draw=GMAxis, line width=1.2pt, fill=GMAxisFill]
        (\tikzlastnode.north west) -- (\tikzlastnode.north east)
        -- ([xshift=2.4mm]\tikzlastnode.east) -- (\tikzlastnode.south east)
        -- (\tikzlastnode.south west) -- ([xshift=-2.4mm]\tikzlastnode.west) -- cycle;
    \end{pgfonlayer}}}},
  indicator/.style={hexbase, text width=33mm, inner ysep=6pt,
    append after command={\pgfextra{\begin{pgfonlayer}{background}
      \path[draw=GMLine, line width=0.7pt, fill=GMFill]
        (\tikzlastnode.north west) -- (\tikzlastnode.north east)
        -- ([xshift=2.4mm]\tikzlastnode.east) -- (\tikzlastnode.south east)
        -- (\tikzlastnode.south west) -- ([xshift=-2.4mm]\tikzlastnode.west) -- cycle;
      \path[draw=GMLine, line width=0.7pt]
        ([yshift=-1.4mm]\tikzlastnode.north west) -- ([yshift=-1.4mm]\tikzlastnode.north east);
      \path[draw=GMLine, line width=0.7pt]
        ([yshift=1.4mm]\tikzlastnode.south west) -- ([yshift=1.4mm]\tikzlastnode.south east);
    \end{pgfonlayer}}}},
  softgoal/.style={
    draw=none, align=center, inner xsep=3pt, inner ysep=13pt,
    text width=30mm, minimum height=13mm,
    append after command={\pgfextra{\begin{pgfonlayer}{background}
      \path[draw=GMLine, line width=0.7pt, fill=GMSoftFill]
        (\tikzlastnode.north west)
          .. controls +(1.2,-0.40) and +(-1.2,-0.40) .. (\tikzlastnode.north east)
          .. controls +(0.5,-0.2) and +(0.5,0.2) .. (\tikzlastnode.south east)
          .. controls +(-1.2,0.40) and +(1.2,0.40) .. (\tikzlastnode.south west)
          .. controls +(-0.5,-0.2) and +(-0.5,0.2) .. cycle;
    \end{pgfonlayer}}}},
  dec/.style={draw=GMLine, line width=0.7pt, 
postaction={
      decorate,
      decoration={
        markings,
        mark=at position 1pt-2mm with {
          \node[transform shape, inner sep=0pt] {
            \tikz\draw[line width=0.7pt] (0,-0.9mm) -- (0,0.9mm);
          };
        }
      }
    }
  },
  contrib/.style={draw=GMLine, line width=0.7pt,
    -{Stealth[length=2mm, width=1.6mm]}},
  op/.style={inner sep=1pt, fill=white, font=\sffamily\scriptsize},
  cw/.style={inner sep=1.5pt, fill=white, font=\sffamily\scriptsize},
]

\node[goal]       (g0) at (-1.6,   0.0) {Every issued fine reaches a lawful, documented closure};
\node[goal]       (g1) at (-4.9,  -1.9) {Fine is issued and formally communicated};
\node[axisroot]   (g2) at ( 1.7,  -1.9) {Fine case is resolved \gmId{4}};
\node[task]       (t1) at (-6.4,  -3.8) {Create Fine};
\node[task]       (t2) at (-3.4,  -3.8) {Send Fine};
\node[anchortask] (tp) at (-0.4,  -3.8) {Resolve via \\ timely payment \gmId{12}};
\node[goal]       (g3) at ( 3.8,  -3.8) {Fine becomes enforceable and is resolved};
\node[task]       (t3) at ( 0.8,  -5.7) {Insert Fine \\ Notification};
\node[task]       (t4) at ( 3.8,  -5.7) {Add Penalty};
\node[goal]       (g4) at ( 6.8,  -5.7) {Enforced case is closed};
\node[anchortask] (ta) at ( 3.4,  -7.6) {Resolve via \\ delinquent \\ payment \gmId{13}};
\node[goal]       (g5) at ( 6.8,  -7.6) {Contested appeal is resolved};
\node[anchortask] (td) at (10.2,  -7.6) {Resolve via \\ coercive credit \\ collection \gmId{20}};
\node[anchortask] (tb) at ( 5.2,  -9.7) {Resolve via \\ administrative \\ appeal \\ (Prefecture) \gmId{14}
                                         \\[-1pt] \gmSub{And: 4 sub-steps}};
\node[anchortask] (tc) at ( 8.4,  -9.7) {Resolve via \\ judicial appeal \\ (Judge) \gmId{19}};
\node[indicator] (kpi) at (-5.6,  -6.6) {Time to fine \\ dispatch (days) \gmId{112}
                                         \\[-1pt] \gmSub{target 30 \textbar{} threshold 90 \textbar{} worst 360}};

\node[softgoal] (sg2) at (-4.6, -12.4) {Maximize timely fine revenue \gmId{22}};
\node[softgoal] (sg3) at ( 6.0, -12.4) {Preserve offender's due-process rights \gmId{23}};
\node[softgoal] (sg1) at (10.4, -12.4) {Minimize administrative \& enforcement cost \gmId{21}};

\begin{pgfonlayer}{background}
  \node[draw=GMLine, line width=0.7pt, dotted, rounded corners=5mm, inner sep=4.5mm,
        fit=(g0)(g1)(g2)(t1)(t2)(tp)(g3)(t3)(t4)(g4)(ta)(g5)(td)(tb)(tc)(kpi)] (actor) {};
\end{pgfonlayer}
\node[anchor=north west, font=\sffamily\small\bfseries, inner sep=2pt, fill=white]
  at ([shift={(3mm,-2mm)}]actor.north west) {Actor: Traffic Police Back-Office};

\draw[dec] (g1.north) -- (g0.south);
\draw[dec] (g2.north) -- (g0.south);
\node[op] at ($(g0.south)+(0.0,-0.34)$) {And};

\draw[dec] (t1.north) -- (g1.south);
\draw[dec] (t2.north) -- (g1.south);
\node[op] at ($(g1.south)+(0.0,-0.34)$) {And};

\draw[dec] (tp.north) -- (g2.south);
\draw[dec] (g3.north) -- (g2.south);
\node[op] at ($(g2.south)+(0.0,-0.34)$) {Or};

\draw[dec] (t3.north) -- (g3.south);
\draw[dec] (t4.north) -- (g3.south);
\draw[dec] (g4.north) -- (g3.south);
\node[op] at ($(g3.south)+(0.0,-0.34)$) {And};

\draw[dec] (ta.north) -- (g4.south);
\draw[dec] (g5.north) -- (g4.south);
\draw[dec] (td.north) -- (g4.south);
\node[op] at ($(g4.south)+(0.0,-0.34)$) {Or};

\draw[dec] (tb.north) -- (g5.south);
\draw[dec] (tc.north) -- (g5.south);
\node[op] at ($(g5.south)+(0.0,-0.34)$) {Xor};

\draw[contrib] (tp.south) to[out=250, in=60]
  node[cw, pos=0.55] {Make (+100)} (sg2.north east);
\draw[contrib] (kpi.south) to[out=285, in=135]
  node[cw, pos=0.5] {Help (+50)} (sg2.north west);
\draw[contrib] (tc.south) to[out=280, in=75]
  node[cw, pos=0.45] {Make (+100)} (sg3.north east);
\draw[contrib] (td.south east) to[out=300, in=80]
  node[cw, pos=0.5] {Hurt ($-$50)} (sg1.north);

\end{tikzpicture}
\endgroup%

%% file: figures/goalcat_pipeline.tikz.tex
%
%

\begingroup%
\definecolor{GCInk}{HTML}{111827}%
\definecolor{GCSlate}{HTML}{475569}%
\definecolor{GCPanel}{HTML}{94A3B8}%
\definecolor{GCHeader}{HTML}{334155}%
\definecolor{GCDetFill}{HTML}{F3F4F6}%
\definecolor{GCBlue}{HTML}{005A8C}%
\definecolor{GCBlueFill}{HTML}{E6F2FA}%
\definecolor{GCOrange}{HTML}{D55E00}%
\definecolor{GCOrangeDark}{HTML}{843800}%
\definecolor{GCOrangeFill}{HTML}{FDF0DD}%
\definecolor{GCGreen}{HTML}{009E73}%
\definecolor{GCGreenDark}{HTML}{005A3C}%
\definecolor{GCGreenFill}{HTML}{E5F5EC}%

\def\gcTag#1{{\fontsize{6.8}{7.5}\selectfont\bfseries\MakeUppercase{#1}}}%
\def\gcTitle#1{{\fontsize{9.2}{10.2}\selectfont\bfseries #1}}%
\def\gcSub#1{{\fontsize{8.0}{8.9}\selectfont #1}}%
\def\gcPhase#1{{\fontsize{9.4}{10.4}\selectfont\bfseries #1}}%
\def\gcEdge#1{{\fontsize{8.0}{8.9}\selectfont #1}}%
\def\gcGate{{\fontsize{10.5}{10.5}\selectfont\bfseries$\times$}}%

\begin{tikzpicture}[
  x=1cm,
  y=1cm,
  font=\sffamily,
  gc/panel/.style={
    draw=GCPanel,
    fill=white,
    line width=0.80pt,
    rounded corners=1.2mm
  },
  gc/phase/.style={
    draw=GCHeader,
    fill=GCHeader,
    text=white,
    align=center,
    minimum height=0.52cm,
    inner xsep=2.2mm,
    inner ysep=1mm,
    rounded corners=0.9mm
  },
  gc/base/.style={
    draw,
    text=GCInk,
    align=center,
    line width=0.65pt,
    inner xsep=1.2mm,
    inner ysep=0.9mm
  },
  gc/deterministic/.style={
    gc/base,
    draw=GCSlate,
    fill=GCDetFill
  },
  gc/llm/.style={
    gc/base,
    draw=GCBlue,
    fill=GCBlueFill,
    rounded corners=1.4mm
  },
  gc/human/.style={
    gc/base,
    draw=GCOrange,
    fill=GCOrangeFill,
    double=white,
    double distance=0.45pt
  },
  gc/artifact/.style={
    gc/base,
    draw=GCSlate,
    fill=white,
    dashed
  },
  gc/output/.style={
    gc/base,
    draw=GCGreen,
    fill=GCGreenFill,
    line width=1.05pt
  },
  gc/gateway/.style={
    diamond,
    aspect=1,
    draw=GCOrange,
    fill=GCOrangeFill,
    text=GCOrangeDark,
    line width=0.90pt,
    minimum width=1.15cm,
    minimum height=1.15cm,
    inner sep=0pt
  },
  gc/gateway merge/.style={
    gc/gateway,
    draw=GCSlate,
    fill=GCDetFill,
    text=GCSlate
  },
  gc/gateway note/.style={
    text=GCSlate,
    align=center,
    inner sep=0.6mm
  },
  gc/flow/.style={
    -{Latex[length=1.7mm,width=1.25mm]},
    draw=GCSlate,
    line width=0.72pt,
    rounded corners=1mm
  },
  gc/data/.style={
    -{Latex[length=1.6mm,width=1.2mm,open]},
    draw=GCSlate,
    line width=0.58pt,
    dash pattern=on 1.7pt off 1.5pt,
    rounded corners=1mm
  },
  gc/revision blue/.style={
    -{Latex[length=1.7mm,width=1.25mm]},
    draw=GCBlue,
    line width=0.85pt,
    rounded corners=1mm
  },
  gc/revision orange/.style={
    -{Latex[length=1.7mm,width=1.25mm]},
    draw=GCOrange,
    line width=0.85pt,
    rounded corners=1mm
  },
  gc/edge label/.style={
    fill=white,
    text=GCSlate,
    inner sep=0.7mm,
    align=center
  }
]

\node[gc/phase, anchor=west] (p1header) at (0.20,1.45)
  {\gcPhase{I. SHARED REPRESENTATION}};

\node[gc/artifact, minimum width=2.70cm, minimum height=1.30cm] (eventlog) at (1.55,0)
  {\textcolor{GCSlate}{\gcTag{Input}}\\[-0.3mm]
   \gcTitle{Event log}\\[-0.3mm]
   \textcolor{GCSlate}{\gcSub{XES + case}}\\[-0.2mm]
   \textcolor{GCSlate}{\gcSub{attributes}}};

\node[gc/deterministic, minimum width=2.70cm, minimum height=1.30cm] (step1) at (4.75,0)
  {\textcolor{GCSlate}{\gcTag{Deterministic}}\\[-0.3mm]
   \gcTitle{1. Extract}\\[-0.2mm]
   \gcTitle{variants}\\[-0.3mm]
   \textcolor{GCSlate}{\gcSub{shared activity}}\\[-0.2mm]
   \textcolor{GCSlate}{\gcSub{sequence}}};

\node[gc/deterministic, minimum width=2.70cm, minimum height=1.30cm] (step2) at (7.95,0)
  {\textcolor{GCSlate}{\gcTag{Deterministic}}\\[-0.3mm]
   \gcTitle{2. Multi-view}\\[-0.2mm]
   \gcTitle{profiles}\\[-0.3mm]
   \textcolor{GCSlate}{\gcSub{order -- time}}\\[-0.2mm]
   \textcolor{GCSlate}{\gcSub{rework -- outcome}}};

\node[gc/deterministic, minimum width=2.70cm, minimum height=1.30cm] (step3) at (11.15,0)
  {\textcolor{GCSlate}{\gcTag{Deterministic}}\\[-0.3mm]
   \gcTitle{3. Variant}\\[-0.2mm]
   \gcTitle{narratives}\\[-0.3mm]
   \textcolor{GCSlate}{\gcSub{one per complete}}\\[-0.2mm]
   \textcolor{GCSlate}{\gcSub{variant}}};

\node[gc/deterministic, minimum width=2.70cm, minimum height=1.30cm] (step4) at (14.35,0)
  {\textcolor{GCSlate}{\gcTag{Deterministic}}\\[-0.3mm]
   \gcTitle{4. Narrative}\\[-0.2mm]
   \gcTitle{sample}\\[-0.3mm]
   \textcolor{GCSlate}{\gcSub{frequent -- rare}}\\[-0.2mm]
   \textcolor{GCSlate}{\gcSub{-- extreme}}};

\node[gc/phase, anchor=west] (p2header) at (0.20,-2.05)
  {\gcPhase{II. MODE-SPECIFIC TAXONOMY}};

\node[gc/gateway] (modegate) at (14.35,-4.80) {\gcGate};
\node[gc/gateway note, anchor=north] (modenote) at (14.4,-5.52)
  {\textcolor{GCOrangeDark}{\gcTag{Mode choice}}\\[-0.3mm]
   \gcSub{\texttt{taxonomy\_mode}}};

\node[gc/human, minimum width=2.45cm, minimum height=1.16cm] (goalmodel) at (11.55,-3.85)
  {\textcolor{GCOrangeDark}{\gcTag{Human}}\\[-0.3mm]
   \gcTitle{G. Goal model}\\[-0.3mm]
   \textcolor{GCSlate}{\gcSub{GRL/URN -- frozen}}\\[-0.2mm]
   \textcolor{GCSlate}{\gcSub{KPI indicators}}};

\node[gc/llm, minimum width=2.45cm, minimum height=1.16cm] (step5a) at (8.80,-3.85)
  {\textcolor{GCBlue}{\gcTag{LLM}}\\[-0.3mm]
   \gcTitle{5a. Guided}\\[-0.2mm]
   \gcTitle{induction}\\[-0.3mm]
   \textcolor{GCSlate}{\gcSub{declared axis}}};

\node[gc/llm, minimum width=2.45cm, minimum height=1.16cm] (step5b) at (8.80,-5.75)
  {\textcolor{GCBlue}{\gcTag{LLM}}\\[-0.3mm]
   \gcTitle{5b. Open}\\[-0.2mm]
   \gcTitle{induction}\\[-0.3mm]
   \textcolor{GCSlate}{\gcSub{sample-derived}}};

\node[gc/artifact, minimum width=2.35cm, minimum height=0.90cm] (anchoredtaxonomy) at (5.95,-3.85)
  {\textcolor{GCSlate}{\gcTag{Artifact}}\\[-0.3mm]
   \gcTitle{Anchored}\\[-0.2mm]
   \gcTitle{taxonomy}};

\node[gc/artifact, minimum width=2.35cm, minimum height=0.90cm] (inducedtaxonomy) at (5.95,-5.75)
  {\textcolor{GCSlate}{\gcTag{Artifact}}\\[-0.3mm]
   \gcTitle{Induced}\\[-0.2mm]
   \gcTitle{taxonomy}};

\node[gc/gateway merge] (mergegate) at (2.60,-4.80) {\gcGate};
\node[gc/gateway note, anchor=south] (mergenote) at (2.60,-4.14)
  {\gcTag{One active lane}};

\node[gc/phase, anchor=west] (p3header) at (0.20,-7.55)
  {\gcPhase{III. CATEGORY ANALYSIS}};

\node[gc/phase, anchor=west] (p4header) at (14.95,-7.55)
  {\gcPhase{IV. VALIDATE AND REPORT}};

\node[gc/artifact, minimum width=2.85cm, minimum height=0.62cm] (assignmentinputs) at (1.55,-8.55)
  {\gcSub{All variant narratives}\\[-0.2mm]\gcSub{(Step 3)}};

\node[gc/llm, minimum width=2.55cm, minimum height=1.30cm] (step6) at (1.55,-10.05)
  {\textcolor{GCBlue}{\gcTag{LLM}}\\[-0.3mm]
   \gcTitle{6. Assign}\\[-0.3mm]
   \gcTitle{narratives}\\[-0.3mm]
   \textcolor{GCSlate}{\gcSub{category, if any}}};

\node[gc/deterministic, minimum width=2.45cm, minimum height=1.30cm] (step7) at (4.45,-10.05)
  {\textcolor{GCSlate}{\gcTag{Deterministic}}\\[-0.3mm]
   \gcTitle{7. Discover}\\[-0.3mm]
   \gcTitle{models}\\[-0.3mm]
   \textcolor{GCSlate}{\gcSub{fitness -- precision}}};

\node[gc/deterministic, minimum width=2.55cm, minimum height=1.30cm] (step7b) at (7.35,-10.05)
  {\textcolor{GCSlate}{\gcTag{Deterministic}}\\[-0.3mm]
   \gcTitle{7b. Indicator}\\[-0.2mm]
   \gcTitle{satisfaction}\\[-0.3mm]
   \textcolor{GCOrangeDark}{\gcSub{guided mode only,}}\\[-0.2mm]
   \textcolor{GCOrangeDark}{\gcSub{optional}}};

\node[gc/deterministic, minimum width=2.40cm, minimum height=1.30cm] (step8a) at (10.20,-10.05)
  {\textcolor{GCSlate}{\gcTag{Deterministic}}\\[-0.3mm]
   \gcTitle{8a. Build}\\[-0.3mm]
   \gcTitle{abstractions}\\[-0.3mm]
   \textcolor{GCSlate}{\gcSub{model -- coverage}}\\[-0.2mm]
   \textcolor{GCSlate}{\gcSub{-- profile}}};

\node[gc/llm, minimum width=2.40cm, minimum height=1.30cm] (step8b) at (12.90,-10.05)
  {\textcolor{GCBlue}{\gcTag{LLM}}\\[-0.3mm]
   \gcTitle{8b. Generate}\\[-0.3mm]
   \gcTitle{descriptions}\\[-0.3mm]
   \textcolor{GCSlate}{\gcSub{one per category}}};

\node[gc/human, minimum width=2.30cm, minimum height=1.30cm] (step9) at (16.30,-10.05)
  {\textcolor{GCOrangeDark}{\gcTag{Human}}\\[-0.3mm]
   \gcTitle{9. Analyst}\\[-0.3mm]
   \gcTitle{review}};

\node[gc/gateway] (reviewgate) at (18.55,-10.05) {\gcGate};

\node[gc/output, minimum width=2.75cm, minimum height=1.30cm] (finaloutput) at (16.30,-12.30)
  {\textcolor{GCGreenDark}{\gcTag{Output}}\\[-0.3mm]
   \gcTitle{Categorization}\\[-0.3mm]
   \gcTitle{outputs}\\[-0.3mm]
   \textcolor{GCSlate}{\gcSub{taxonomy -- log}}\\[-0.2mm]
   \textcolor{GCSlate}{\gcSub{coverage report}}};

\node[gc/artifact, minimum width=6.90cm, minimum height=1.22cm] (residual) at (4.10,-12.15)
  {\textcolor{GCSlate}{\gcTag{Unassigned narratives -- residual}}\\[-0.3mm]
   \gcSub{Guided mode: candidate goal-model coverage gap}\\[-0.2mm]
   \gcSub{Open mode: other / unclassifiable}};

\node[gc/artifact, minimum width=5.40cm, minimum height=1.05cm] (measured) at (10.85,-12.15)
  {\textcolor{GCSlate}{\gcTag{Measured goal satisfaction}}\\[-0.3mm]
   \gcSub{Indicator + goal satisfaction (CSV)}\\[-0.2mm]
   \gcSub{Measured goal model (.jucm)}};

\coordinate (p4south) at (16.30,-13.42);

\begin{pgfonlayer}{background}
  \node[gc/panel, fit=(p1header)(eventlog)(step4), inner xsep=2.2mm, inner ysep=2.1mm] {};
  \node[gc/panel, fit=(p2header)(modegate)(modenote)(goalmodel)(step5a)(step5b)(anchoredtaxonomy)(inducedtaxonomy)(mergegate)(mergenote), inner xsep=2.2mm, inner ysep=2.1mm] {};
  \node[gc/panel, fit=(p3header)(assignmentinputs)(step6)(step8b)(residual)(measured), inner xsep=2.2mm, inner ysep=2.1mm] {};
  \node[gc/panel, fit=(p4header)(step9)(reviewgate)(finaloutput)(p4south), inner xsep=2.2mm, inner ysep=2.1mm] {};
\end{pgfonlayer}

\draw[gc/flow] (eventlog) -- (step1);
\draw[gc/flow] (step1) -- (step2);
\draw[gc/flow] (step2) -- (step3);
\draw[gc/flow] (step3) -- (step4);

\draw[gc/flow] (step4.south) --
  node[pos=0.42, gc/edge label]{\gcEdge{sample}}
  (modegate.north);

\draw[gc/flow] (modegate.west) -- (13.40,-4.80) --
  node[pos=0.55, gc/edge label]{\gcEdge{guided}}
  (13.40,-3.85) -- (goalmodel.east);
\draw[gc/flow] (modegate.west) -- (13.40,-4.80) -- (13.40,-5.75) --
  node[pos=0.32, gc/edge label, text=GCSlate]{\gcEdge{\texttt{open}}\\\gcEdge{(no goal model)}}
  (step5b.east);

\draw[gc/flow] (goalmodel) -- (step5a);
\draw[gc/flow] (step5a) -- (anchoredtaxonomy);
\draw[gc/flow] (step5b) -- (inducedtaxonomy);

\draw[gc/flow] (anchoredtaxonomy.west) -- (3.85,-3.85) -- (3.85,-4.80) -- (mergegate.east);
\draw[gc/flow] (inducedtaxonomy.west) -- (3.85,-5.75) -- (3.85,-4.80) -- (mergegate.east);

\draw[gc/flow] (mergegate.west) -- (-0.30,-4.80) --
  node[pos=0.55, gc/edge label, rotate=90]{\gcEdge{selected taxonomy}}
  (-0.30,-10.05) -- (step6.west);

\draw[gc/data] (assignmentinputs) -- (step6);

\draw[gc/flow] (step6) -- (step7);
\draw[gc/flow] (step7) -- (step7b);
\draw[gc/flow] (step7b) -- (step8a);
\draw[gc/flow] (step8a) -- (step8b);
\draw[gc/flow] (step8b) -- (step9);
\draw[gc/flow] (step9) -- (reviewgate);
\draw[gc/flow] (reviewgate.south) --
  node[pos=0.25, gc/edge label, anchor=east, xshift=11mm]{\gcEdge{accept / rename}}
  (18.55,-12.30) -- (finaloutput.east);

\draw[gc/data] (step6.south) --
  node[gc/edge label]{\gcEdge{unassigned}}
  ([xshift=-2.55cm]residual.north);
\draw[gc/data] (residual.south) -- (4.10,-13.2) --
  node[pos=0.18, gc/edge label]{\gcEdge{reported}}
  ([yshift=-8.9mm]finaloutput.west);

\draw[gc/data] (step7b.south) -- (7.35,-11.25) -- (8.65,-11.25) --
  ([xshift=-2.2cm]measured.north);
\draw[gc/data] (measured.east) -- ([yshift=1.5mm]finaloutput.west);

\draw[gc/data] ([xshift=-3.5mm]goalmodel.south) -- (11.20,-8.30) -- (7.05,-8.30) --
  ([xshift=-3mm]step7b.north);
\draw[gc/data] (11.20,-8.30) --
  node[pos=0.50, gc/edge label, anchor=east, xshift=-1mm]{\gcEdge{KPI indicators (guided mode only)}}
  (13.50,-8.30) -- ([xshift=6mm]step8b.north);
\fill[GCSlate] (11.20,-8.30) circle (0.65pt);

\draw[gc/data] ([xshift=6mm]step7b.north) -- (7.95,-8.85) --
  node[pos=0.5, gc/edge label]{\gcEdge{measured satisfaction, if present}}
  (12.30,-8.85) -- ([xshift=-6mm]step8b.north);

\draw[gc/flow] (reviewgate.north) -- (18.55,-9) -- (20.20,-9) -- (20.20,-4.80) --
  node[pos=0.41, gc/edge label, align=center]
    {\gcEdge{merge / split -- repeat 5--9}\\\gcEdge{(rename: in place, no LLM call)}}
  (modegate.east);

\draw[gc/revision orange] (reviewgate.east) -- (20.50,-10.05) -- (20.50,-2.75) --
  node[pos=0.28, gc/edge label, text=GCOrangeDark]{\gcEdge{guided mode only -- revise goal model}}
  (11.55,-2.75) -- (goalmodel.north);

\end{tikzpicture}
\endgroup